\documentclass[journal]{IEEEtran}

\usepackage{cite}
\usepackage{amsmath}
\usepackage[english]{babel}
\usepackage[export]{adjustbox}

\begin{document}
%
\title{Optimal Scheduling of Electrolyzer in Power Market with Dynamic Prices}
%
%
%
\author{\IEEEauthorblockN{Yusheng Luo\IEEEauthorrefmark{1},
Min Xian\IEEEauthorrefmark{2},
Manish Mohanpurkar\IEEEauthorrefmark{1},
Bishnu P. Bhattarai\IEEEauthorrefmark{1},
Anudeep Medam\IEEEauthorrefmark{1},
Rahul Kadavil\IEEEauthorrefmark{1} and\\
Rob Hovsapian\IEEEauthorrefmark{1}}\\
\IEEEauthorblockA{\IEEEauthorrefmark{1}Idaho National Laboratory\\
Idaho Falls, Idaho 83415\\ Email: {yusheng.luo, Manish.Mohanpurkar, Bishnu.Bhattarai, Anudeep.Medam, Rahul.Kadavil, Rob.Hovsapian}@inl.gov\\}
\IEEEauthorblockA{\IEEEauthorrefmark{2}University of Idaho\\
Idaho Falls, Idaho 83402\\
Email: mxian@uidaho.edu}}

\maketitle

\begin{abstract}
Optimal scheduling of hydrogen production in dynamic pricing power market can maximize the profit of hydrogen producer; however, it highly depends on the accurate forecast of hydrogen consumption. In this paper, we propose a deep leaning based forecasting approach for predicting hydrogen consumption of fuel cell vehicles in future taxi industry. The cost of hydrogen production is minimized by utilizing the proposed forecasting tool to reduce the hydrogen produced during high cost on-peak hours and guide hydrogen producer to store sufficient hydrogen during low cost off-peak hours.
\end{abstract}


\IEEEpeerreviewmaketitle

\section{Introduction}
\IEEEPARstart{F}{uel} Cell Vehicle (FCV) powered by hydrogen is a promising transportation alternative in the future due to the great advantages of being extremely clean\cite{fcevmost}, highly efficient and scalable\cite{wei2014}. Hydrogen, one of the most abundant elements in the earth, is regarded as the cleanest fuel that has great potential to replace the carbon-based fuels. Consequently, highly-economic production of hydrogen becomes crucial to the hydrogen economy. At present, most hydrogen is produced from steam reforming of fossil fuels (steam methane reforming, SMR), which is a mature technique due to its operational reliability and low cost of methane since the shale gas revolution. However, the fossil fuels are not renewable. Nevertheless the exhausts (e.g., CO2 and SO2) are threats to the environment. Therefore it has to be acknowledged that SMR is neither sustainable nor eco-friendly. Most importantly, SMR has low cost-scaling factor of a conventional reformer, so it is a relatively expensive option for distributed or small-scale applications. Consequently, steam electrolysis is a viable alternative to SMR in specific applications owing to its very high cost scaling factors and clean products.

With the breakthrough in electrolyzer technology, effectively producing hydrogen using distributed electrolyzer becomes feasible \cite{mohanpurkar2017electrolyzers}. When electricity energy can steadily power electrolyzer, FCV based transportation systems can significantly reduce impact of human travel to environment.

However, power system generating electricity to electrolyzer cannot always guarantee stable power supply. When total rated power of electrical load plugged into the power grid of an area reaches or exceeds the upper limit of power supply capacity from utility, power quality metrics such as voltage or frequency can be seriously affected. This trend can in turn affect the performance of loads. In order to keep load demand within the limit of power supply during peak hour when power deficit is most likely to happen, utilities set high price of electricity during peak hour and/or forcibly cut the power supply to some users when power quality metrics further deviate from expected value. Needless to say, excessively consuming large amount of electricity power during peak time can affect both power grid operation and backlash load itself. Even the reaction from grid cause little damage to device, the bill can be much larger than that charged to the same amount of energy during off peak time.

Utilities take dynamic pricing strategy to encourage the most efficient use of electricity and reduce the cost of electricity. According to the statistical analysis of power system data, regional power consumption demonstrates a \emph{spatiotemporal} characteristic. Based on this feature, one typical approach is to divide the bill rate into two levels : on-peak bill rate and off-peak rate\cite{centolella2010integration}. During peak hours, consumption of electricity power is billed at a high rate, \textit{i.e. }more than ten cents per kilowatt-hour due to the reason that the cost of procuring electricity and managing grid-operation is much higher than that in any other time of the day\cite{coned}. While during the off-peak hours, the bill rate can be as low as only one cent. With this invisible hand, utilities are able to encourage a shifting of a considerable portion of peak time electricity usage to off-peak hours\cite{herter2007residential}.

Unplanned hydrogen production could lead to very high cost due to the implementation of dynamic pricing. Therefore, it is vital important to find an optimal scheduling approach that can minimize the cost for electricity. Electrical water heaters have been embedded with night mode, in which boiling hot water is mainly performed during cheap off-peak time. This strategy also works for products which can be stored before selling them. Production of hydrogen happens to fall into this category. 

Currently, hydrogen production is still in a spontaneous mode. Hydrogen consumption forecasting is neglected due to lack of forecasting method. This can either give rise to producing too much hydrogen during high price period or experiencing hydrogen shortage.

Deep Learning-based approaches have been reported to achieve state-of-the-art performance for many big data tasks such as object recognition \cite{cirecsan_NN_2012}, image classification\cite{ciregan_CVPR_2012}, speech recognition \cite{deng_ICASSP_2013}, medical applications \cite{kooi_MIA_2017,cheng_Science_2016}, etc. Since power system engineers have started to analyze mass data sampled from power system operation and gain incisive understanding of the operation status\cite{huaiguangmost,jiang2016spatial}, deep learning technology is deemed as the one of the key factors for grid modernization.

In this work, we propose a real-time deep learning-based approach to estimate the total hydrogen consumption of FCV taxi in a future urban area. The approach can contribute to the goal of minimizing cost of hydrogen production by applying an optimal production scheduling based on accurate and real-time consumption forecasting.

\section{Background}
We utilize a city model built with the transportation data of New York City (NYC). These data are shared by NYC Taxi \& Limousine Commission (NYCTLC)\cite{nyctrip} and \cite{nyctaxi} and has already facilitated many research \cite{barbosa2014structured,castellani2015urban,poco2015exploring}.

Taxi industry aggressively seeks fuel economy. Due to the high Miles Per Gallon-equivalent (MPGe) of FCV, it has great potential that in the near future FCV could be the best option to  undertake the transportation task in most cities of the world. Besides the convenience of accessing New York Taxi data, the most important reason of choosing NYC is the sharp  difference (20 times) of electricity prices between peak and off-peak periods in the time-of-use program of Con-Edison (utility of NYC). If the hydrogen production is not scheduled wisely, e.g., overproduction during peak hours, the average electricity cost of hydrogen production could be very high, which weakens the advantage of FCV. If most of the hydrogen sold during a day is produced in off-peak hour, the profit of hydrogen refueling station can soar, vice versa. 

\begin{table}
\centering
\caption{Consolidated Edison Time-of-use bill rate for business customers}
\label{nycbill}
\begin{tabular}{|l|l|l|}
\hline
 June 1 to Sept 30&27.61 cents/kWh  &1.01 cents/kWh  \\ \hline
 All other months&13.60 cents/kWh  &1.01 cents/kWh  \\ \hline
\end{tabular}
\end{table}

We use the transportation data from NYCTLC to model the New York Taxi market and assume all of taxis in the studied model are FCV. Based on the data, the real time hydrogen consumption is predicted by using a well-trained deep learning model, so that during off-peak hours electrolyzer can fill the tank with hydrogen, while produce minimum required hydrogen during peak hours.

In each economic optimization period, which is between $T_K$ and $T_{k+1}$, the objective is to generate as much hydrogen as possible and subject to 
\begin{equation}
\label{slow1}
H_k+GE_k-W_k\leq H_{max}, and 
\end{equation}
\begin{equation*}
E_k\leq E_{max}
\end{equation*}
\\where $H_k$ is the amount of hydrogen stored at the beginning of this optimization cycle;  $G$ is the conversion rate between hydrogen production and electric power $E_k$ in each period; we assume that G is constant for all value of $E_k$; $W_k$ is the current forecasting of hydrogen demand; $H_{max}$ is the maximum allowed hydrogen storage; and $E_{max}$ is the maximum allowed electricity consumption during each optimization cycle. 

Following the strategy of maximizing hydrogen production during off-peak hours, the maximum electricity consumption is given by
\begin{equation}
\label{slow2}
E_k =min(E_{max},\frac{H_{max}+W_k-H_k}{G}).
\end{equation}

During peak hours, the objective is to produce as less hydrogen as possible while the hydrogen storage should never be empty with the following constraint
\begin{equation}
\label{slow3}
H_k+GE_k-W_k\geq H_{min}
\end{equation}
\\where $H_{min}$ is the required minimum level of hydrogen storage and introduced to cater to an large unexpected increase of hydrogen demand. The minimum amount of electricity consumption during each optimization period is defined by
\begin{equation}
\label{slow4}
E_k =max (0,\frac{H_{min}+W_k-H_k}{G}).
\end{equation}

\begin{figure}[!ht]
\centering
\includegraphics[width=\linewidth]{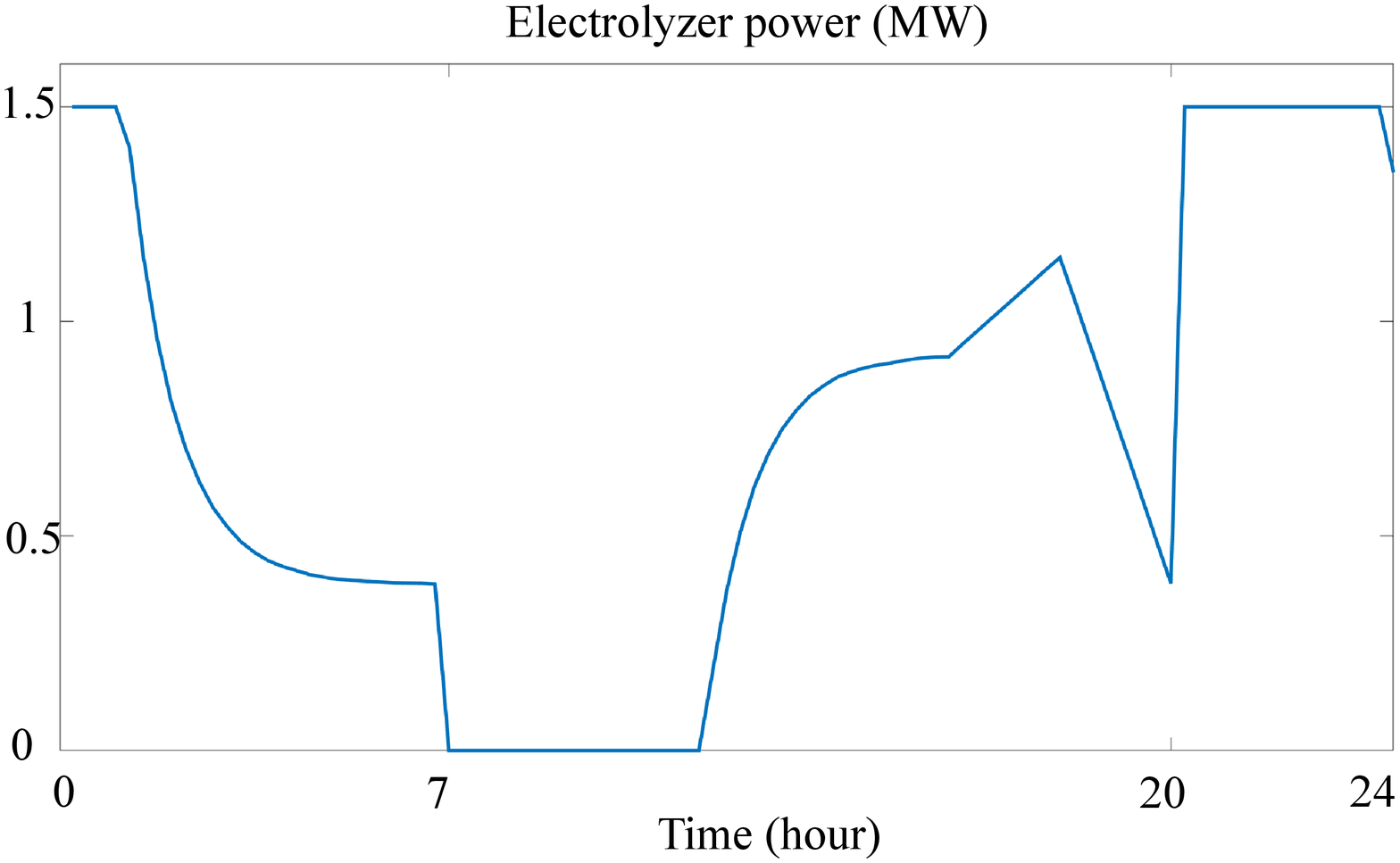}
\caption{Optimal power consumption of electrolyzer based on accurate demand forecasting.}
\label{pact}
\end{figure}

\begin{figure}[!ht]
\centering
\includegraphics[width=\linewidth]{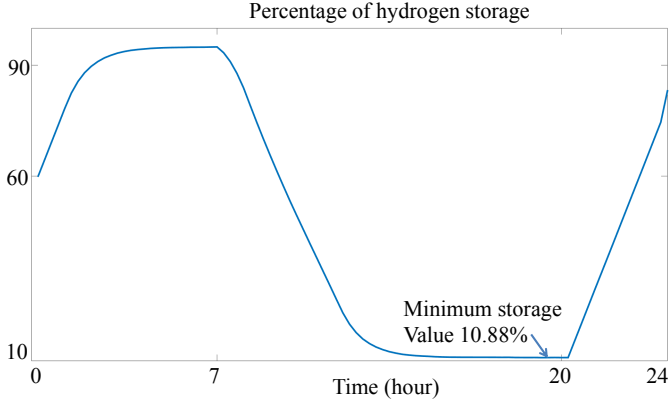}
\caption{Hydrogen storage change in accurate demand forecasting case.}
\label{h2store}
\end{figure}

As shown in Eqs. (\ref{slow2}) and (\ref{slow4}), the accurate forecasting of hydrogen consumption plays a critical role in minimizing hydrogen production cost. In this case, peak hours starts from 7 o'clock and ends at 20 o'clock. The minimum required storage value is set at 10\%. The conversion rate between power and hydrogen is set as 55 kW/kg. Figure~\ref{pact} shows the optimal scheduling of hydrogen production according to the accurate forecasting of hydrogen consumption. Figure~\ref{h2store} demonstrates that, under this optimal control, hydrogen storage in tank can be maintained above the preset minimum value.

In order to provide guidance for economic hydrogen production, we initiate a study on applying deep learning to forecast hydrogen consumption of NYC in real-time. In this paper, we aim at the first stage forecasting: precisely estimate the total hydrogen consumption of all FCV taxis based on the analysis on existing taxi driving data.

\section{Proposed Approach}
\subsection{Overview}
Accurate and real-time forecast of the hydrogen consumption is the key to plan the production and lower the total cost of electricity consumption. We utilize the NYC yellow taxi dataset to train a deep learning model to forecast, in every 15 minutes, the pounds of hydrogen will be consumed next hour. 
The consumption of hydrogen in the \textit{k}th optimization period is given by
\begin{equation}
\label{h2con}
H_k=\frac{D_k}{mpge}=\frac{\bar{v}_k\cdot n_k}{mpge}
\end{equation}
\\where the unit of $H_k$ is pound, $D_k$ denotes the total driving distances of all taxis, $\bar{v_k}$ is the average traffic speed, $n_k$ is the number of taxis on road, and mpge is a constant and denotes the average miles a taxi can travel using 1 pound hydrogen. 

Let $TD_k$  be the total recorded trip distances in kth optimization, and $TT_k$ be the corresponding trips hours; and the average speed is defined by
\begin{equation}
\label{avv}
\bar{v_t}=\frac{TD_k}{TT_k}
\end{equation}
\\where $n_k$ is not directly given by the data set and we assume that it has the following linear relation with total number of trips $nt_k$
\begin{equation}
\label{nk}
n_k=f(nt_k)=a\cdot nt_k+b,
\end{equation}
\\where $a$ and $b$ are estimation by using linear regression.

According to Eqs.(\ref{h2con})-(\ref{nk}), we know exactly how many pounds of hydrogen are consumed if we have $\bar{v}_k$ and $n_k$; however, in practice,  $\bar{v}_k$ and $n_k$  are not available, and consequently, we cannot simply apply these equations to forecast the future hydrogen consumption. 

\begin{figure}[!ht]
\centering
\includegraphics[width=\linewidth]{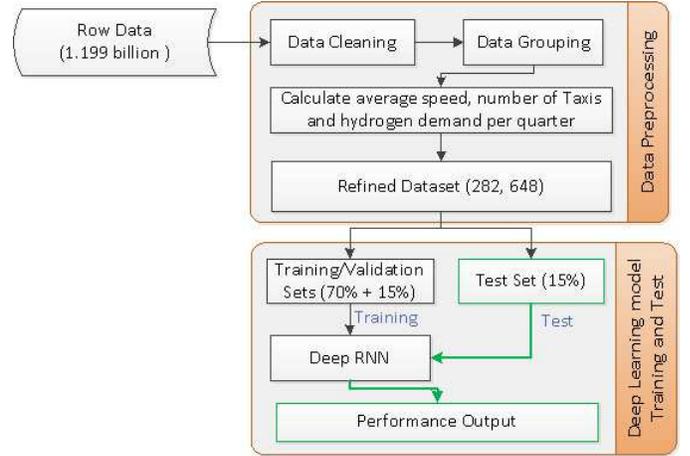}
\caption{Flowchart of the proposed approach.}
\label{flowchart}
\end{figure}

In the \textit{k}th optimization period, two data sequences $(\bar{v}_0,\bar{v}_1,\ldots,\bar{v}_{k-1})$ and $(n_0,n_1,\ldots,n_{k-1})$ of the previous time steps could be available; and our main idea is to forecast the hydrogen consumption by training a deep neural network using the historical sequential data. The flowchart of the newly proposed approach is shown in Figure~\ref{flowchart}.

\subsection{Data Cleaning and Grouping}
We use the monthly reports and daily trips data of the yellow taxis in New York City from January 2010 to June 2017. The daily trip data recorded detail information of every trip, e.g., date, pick time, drop off time, distance, etc., but large amount of dirty data. The 1.199 billion trip data records are cleaned based on the three rules
\begin{equation*}
\label{rule1}
tripDis_i>0 (miles),
\end{equation*}
\begin{equation*}
\label{rule2}
\frac{1}{60}<rideTime_i< 3 (hours),
\end{equation*}
\begin{equation*}
\label{rule3}
and \ 0<\frac{tripDis_i}{rideTime_i}\leq 80 (miles\ per\ hour)
\end{equation*}

In the three rules, $tripDis_i$ and $rideTime_i$ are the trip distance and the ride time of the \textit{i}th record, respectively. The cleaning removes 22.8 million dirty records that accounts for 1.9\% of the original data.
In the grouping step, all trip data in a quarter is grouped together to generate one record which contains the corresponding date, hour (0 - 23), quarter indicator (1 - 4), sum of trip distances, sum of ride time, total trip numbers.

\subsection{Hydrogen Consumption Calculation}
As shown in Eq. (\ref{h2con}), to calculate the quarterly hydrogen consumption, we should have the average speed, number of operating taxis of that quarter hour. The average speed could be calculated easily by using Eq. (\ref{avv}); because the number of operating taxis are not available, we estimate it indirectly by using the number of trips and monthly total operating hours of all taxis from the monthly trip report.
Suppose that a taxi operating at operation period k will run the full corresponding hour; and we can have the following relation based on Eq. (\ref{nk})

\begin{figure}[!ht]
\centering
\includegraphics[width=\linewidth]{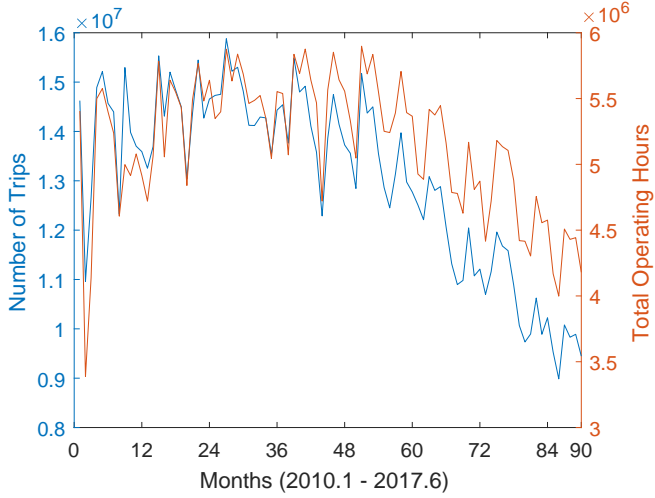}
\caption{Monthly total number of trips and operation hours.}
\label{method}
\end{figure}

\begin{equation}
\label{relation}
MOH_m=\sum_{d=1}^{MD} \sum_{h=1}^{24} n_{d,h}^m=a\cdot \sum_{d=1}^{MD}\sum_{h=1}^{24}nt_{d,h}^m +MD\cdot24\cdot b.
\end{equation}

To simplify the equation, we set $b_1=MD\cdot24\cdot b$, therefore, we have
\begin{equation}
\label{relationsimp}
MOH_m=a\cdot \sum_{d=1}^{MD}\sum_{h=1}^{24}nt_{d,h}^m +b_1.
\end{equation}
\\where $MD$ is the number of days of the $m$ month and take integer values from $[0,90]$, $n_{d,h}^m$ and $nt_{d,h}^m$ are the number of operating taxis and the number of total trips from h to $h+1$ o’clock of the dth day of the mth month. With the values of $a$ and $b_1$ , the hourly hydrogen demand could be computed by using Eqs. (\ref{h2con})--(\ref{nk}).
\subsection{Deep Recurrent Neural Network}
\begin{figure}[!ht]
\centering
\includegraphics[width=\linewidth]{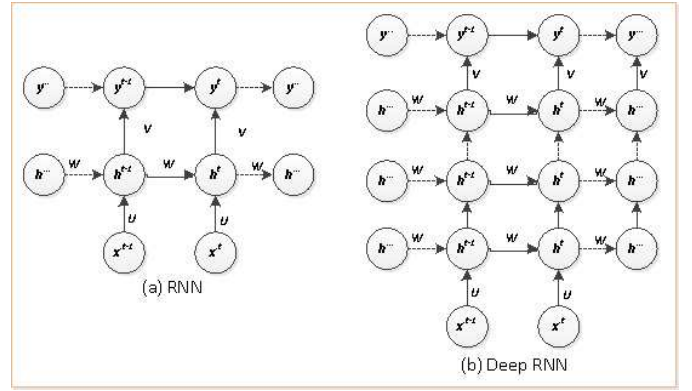}
\caption{RNN and deep RNN.}
\label{rnn}
\end{figure}

The above section discussed the approach to calculate the hydrogen demand by using the average speed and number of operating taxis at time step $k$. In this section, we train a deep learning model to forecast the hydrogen demand at time step $k+1$ by using historical data from time step 1 to $k$. Let $x^1$,$x^2$,\ldots,$x^k$  be the input time series where $x^k$ denotes the hydrogen demand between time $k$ and $k+1$; our task is to forecast $x^k$ based on the observation of $x^1$,$x^2$,\ldots,$x^k-1$; for clarity we will utilize $y^1$, $y^2$,\ldots,$y^k$ to denote the output sequence.
A recurrent neural network (RNN) is neural network that models a dynamic system that has an input  $x^k$, ouput $y^k$ and hidden state $h^k$. Our preference of RNN for this work is because 1) RNN is good at model the internal dependence of data sequence; 2) RNN shares parameter at all time steps and thus has much less parameters than other neural nets; and 3) use feedback connections to introduce historical information. Three are three design patterns of RNNs: 1) having recurrent connections between hidden node; produces an output at each time step; 2) only having recurrent connection from the output at one time step to the hidden nodes at the next time step; 3) reading entire sequence to produce on output. In this paper we utilize and implement the second pattern of RNN and the structure is shown is Figure~\ref{rnn}(a).
In Figure~\ref{rnn}, $W$ is a weight matrix that defines the parameters of the hidden-to-hidden connections, $U$ is weight matrix of the input-to-hidden connections, and $V$ denotes the hidden-to-output weight matrix. For the hidden unit, we specify the hyperbolic tangent activation function. We apply the following equations for each time step k between 1 and $\tau$.

\begin{equation}
\label{hk}
h^k=tanh(a^k),
\end{equation}

\begin{equation}
\label{ak}
a^k=Wh^k+Ux^k+c_1,
\end{equation}

\begin{equation}
\label{yk}
\hat{y}^k=Vh^k+c_2.
\end{equation}

In Eqs.(\ref{ak}) and (\ref{yk}), $c_1$ and $c_2$ are bias vectors for input-to-hidden and hidden-to-output connections, respectively. The loss function is defined on the target sequence $y^k$ and the forecasted sequence $\hat{y}^k$:

\begin{equation}
\label{loss}
L(y^k,\hat{y}^k)=\sum_{t=1}^\tau d(y^t,\hat{y}^k).
\end{equation}
\\where $d$ defines the cross-entropy between two $y^k$ and $\hat{y}^k$.

Eqs.(\ref{hk})--(\ref{loss}) define a typical shallow RNN; however, many researchers have demonstrated that a deep, hierarchical RNN could be much more efficient and generate better results, which leads to increase the depth of RNN by employing multiple nonlinear layers between the input and output layers. In this work, we make the hidden-to-hidden layers deeper by introducing more intermediate nonlinear hidden layers between the hidden layer to the output layer because this structure has the advantage of summarizing the history of previous input efficiently. The structure of the deep RNN is shown in Figure~\ref{rnn}(b). The parameters $W$, $U$, $b_1$, and $c$ are estimated by using the backpropagation algorithm during the training step.

\section{Experimental results}
\subsection{Dataset and Parameters}
We utilize the Yellow-Taxi trip dataset of NYC from January 2010 to June 2017. It has 1.199 billion trip records. After the data preprocessing step, a clean dataset with 262,848 entries is generated, and each entry contains the month, day, day of week, the number of trips, the number of cars and average speed in one period of 15 minutes.

The MPGe is set to 30 miles per pound of hydrogen; the parameters $a$ and $b_1$ are estimated in section III(c), respectively 0.247 and $2\times 10^6$. The deep RNN has one input layer, one output layer, and five hidden layers; and each hidden layer has five hidden units. All the parameters, $W$, $U$, $V$, $c_1$, and $c_2$, are estimated by using the training sets.

\subsection{Hydrogen Consumption Calculation}
\begin{figure}[!ht]
\centering
\includegraphics[width=\linewidth]{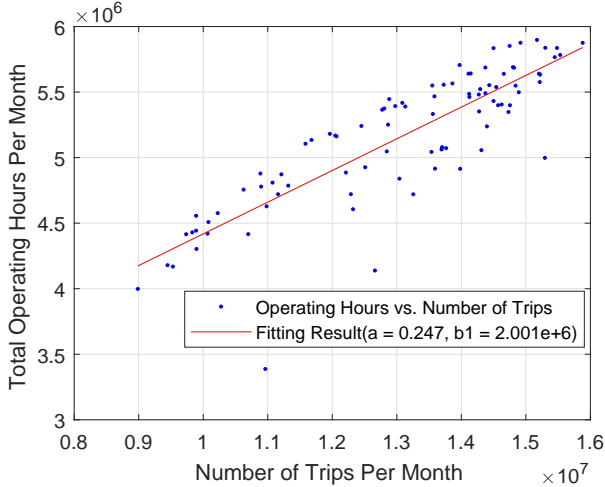}
\caption{Linear regression result.}
\label{rnnresult}
\end{figure}

\begin{figure}[!ht]
\centering
\includegraphics[width=\linewidth]{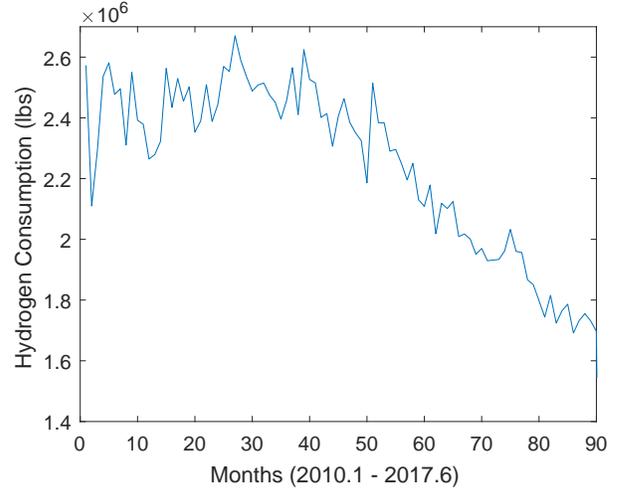}
\caption{Monthly hydrogen demand.}
\label{mondemand}
\end{figure}

The two parameters $a$ and $b_1$ in Eq. (\ref{relationsimp}) are estimated by using linear regression and the result is demonstrated in Figure~\ref{rnnresult}.  An apparent outlier (in pink circle) exists in Figure~\ref{rnnresult}, and shows quite low total operating hours of the second month of 2010. Acceding to the monthly report, this is caused by a great shrink of the number of average operating yellow taxis per day (2010.01: 12, 727 taxis, 2010.02: 10, 045 taxis) and much less average operating days per taxi (2010.01: 29.7 days, 2010.02: 23.1 days). 
Figure~\ref{mondemand} shows the monthly hydrogen consumption of the yellow taxis calculated by using Eq. (\ref{relationsimp}), and demonstrated a steady downward trend after 2013.  The emerging travel options, e.g., Uber and Green Taxis (2013), could be the major driving forces of the trend.

\subsection{Hydrogen Consumption Forecasting}
\begin{figure}[!ht]
\centering
\includegraphics[width=\linewidth]{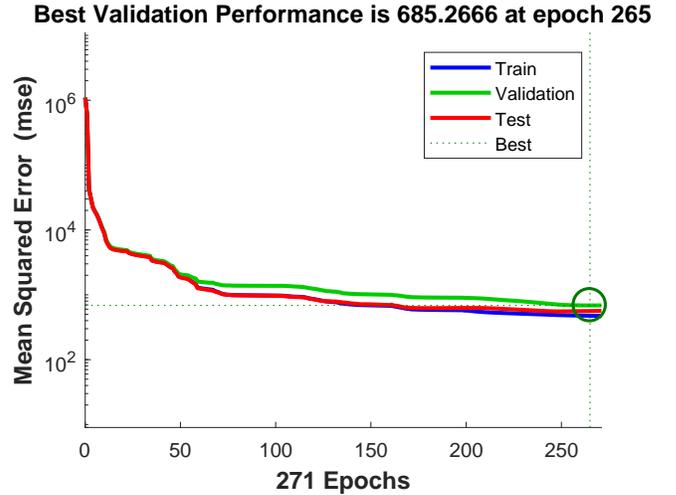}
\caption{Performance evaluation..}
\label{pandv}
\end{figure}

To evaluate the performance of the proposed approach, we randomly divide the dataset into training (70\%), validation (15\%) and test (15\%) sets; the training and validation sets are applied to learn the best parameters of the deep nets; and the training process lasts 8.5 hours with 271 iterations. In Figure~\ref{pandv}, the horizontal axis is the number of iterations, and the vertical axis represents the mean square error (mse) between the target and the forecasting; the red, green and blue curves respectively demonstrate the model performances on the training, validation and test sets. The best performance is achieved at the 265th iteration. 

\section{Conclusion}
In this work, we focus on reducing the cost of producing hydrogen by accurately forecasting the hourly hydrogen consumption of  NYC\'s yellow taxis. The proposed approach follows a two-step strategy: data preprocessing and deep RNN training and test. The data preprocessing step calculates the average taxis speed, and estimates the total number of operating taxis by utilizing linear regressing. In the second step, a deep RNN is trained to forecast the hourly hydrogen consumption in the future based on historical data.

The proposed deep RNN-based approach achieve accurate and real-time forecast due to the following reasons: 1) the linear regression accurately estimated the relationship between the total number of yellow taxis and the number of trip records; and 2) the seven-layer deep RNN can learn complicated internal relations among sequential data. In the future, we will estimate the hydrogen consumption of different locations in NYC, which could be very useful to plan hydrogen refueling stations and their corresponding capacities.

\section*{Acknowledgment}

Funding support from the Fuel Cell Technologies Office, Department of Energy is acknowledged.

\ifCLASSOPTIONcaptionsoff
  \newpage
\fi


\bibliographystyle{IEEEtran}
\bibliography{IEEEexample}
\end{document}